\documentclass[11pt]{article}
\usepackage[utf8]{inputenc}
\usepackage{hyperref}
\title{Neurosymbolic AI: The $3^{rd}$ Wave}
\author{\textsc{Artur d'Avila Garcez$^1$ and Luís C. Lamb$^2$}\\ 
$^1$ City, University of London, UK\\
a.garcez@city.ac.uk\\ 
$^2$ Federal University of Rio Grande do Sul, Brazil\\
luislamb@acm.org}
\date{December, 2020}

\begin{document}
\maketitle
\begin{abstract}
Current advances in Artificial Intelligence (AI) and Machine Learning (ML) have achieved unprecedented impact across research communities and industry. Nevertheless, concerns about trust, safety, interpretability and accountability of AI were raised by influential thinkers. Many have identified the need for well-founded knowledge representation and reasoning to be integrated with deep learning and for sound explainability. Neural-symbolic computing has been an active area of research for many years seeking to bring together robust learning in neural networks with reasoning and explainability via symbolic representations for network models. In this paper, we relate recent and early research results in neurosymbolic AI with the objective of identifying the key ingredients of the next wave of AI systems. We focus on research that integrates in a principled way neural network-based learning with symbolic knowledge representation and logical reasoning. The insights provided by 20 years of neural-symbolic computing are shown to shed new light onto the increasingly prominent role of trust, safety, interpretability and accountability of AI. We also identify promising directions and challenges for the next decade of AI research from the perspective of neural-symbolic systems.\\ 
\end{abstract}


\section{Introduction}

Over the past decade, Artificial Intelligence and in particular deep learning have attracted media attention, have become the focus of increasingly large research endeavours, and have changed businesses. This led to influential debates on the impact of AI both on academia and industry 
\cite{marcus2020}, \cite{raghavan19}. It has been claimed that Deep Learning (DL) caused a paradigm shift not only in AI, but in several Computer Science fields, including speech recognition, computer vision and image understanding, natural language processing (NLP) and machine translation \cite{Hinton-nature}. 
The 2019 Montréal AI Debate between Yoshua Bengio and Gary Marcus, mediated by Vincent Boucher \cite{marcus2020}, and the AAAI-2020 fireside conversation with Economics Nobel Laureate Daniel Kahneman, mediated by Francesca Rossi and including the 2018 Turing Award winners and DL pioneers Geoffrey Hinton, Yoshua Bengio and Yann LeCun, have pointed to new perspectives and concerns on the future of AI. It has now been argued eloquently that if the aim is to build a rich AI system, that is, a semantically sound, explainable and ultimately trustworthy AI system, one needs to include with it a sound reasoning layer in combination with deep learning. 
Kahneman corroborated this point at AAAI-2020 by stating that \emph{...as far as I'm concerned, System 1 certainly knows language... System 2 does involve certain manipulation of symbols} \cite{fireside2020}. Kahneman's comments at AAAI-2020 go to the heart of the matter, with parallels having been drawn many times by AI researchers between Kahneman's research on human reasoning and decision making - reflected in his book ``Thinking, Fast and Slow" \cite{Kahneman2011} - and the so-called ``AI systems 1 and 2", which would in principle be modelled by deep learning and symbolic reasoning, respectively.\footnote{``Thinking Fast and Slow", by Daniel Kahneman: New York: Farrar, Straus and Giroux, 2011, describes the author's \emph{``... current understanding of judgement and decision making, which has been shaped by psychological discoveries of recent decades."} Of course, the concepts of systems 1 and 2 derive from decades of research in Psychology and Cognitive Science and a comprehensive explanation is beyond the scope and aims of this paper.}   

In this paper, we place 20 years of research from the area of neurosymbolic AI, known as neural-symbolic integration, in the context of the recent explosion of interest and excitement about the combination of deep learning and symbolic reasoning. We revisit early theoretical results of fundamental relevance to shaping the latest research, and identify bottlenecks and the most promising technical directions for the sound representation of learning and reasoning in neural and symbolic systems. 

As well as pointing to the various related and promising techniques within AI, ML and Deep Learning, this article seeks to help organise some of the terminology commonly used around AI. This seems important at this exciting time when AI becomes popularized and more people from other areas of Computer Science and from other fields altogether turn to AI: psychology, cognitive science, economics, medicine, engineering and neuroscience to name a few.

In Section 2, we position the current debate in the context of the necessary and sufficient building blocks of AI and long-standing challenges of variable grounding and commonsense reasoning. In Section 3, we seek to organise the debate, which can become vague if defined around the concepts of \emph{neurons versus symbols}, around the concepts of distributed and localist representations. We argue for the importance of this focus on representation since representation precedes learning as well as reasoning. We also analyse a taxonomy for neurosymbolic AI proposed by Henry Kautz at AAAI-2020 from the angle of localist and distributed representations. In Section 4, we delve deeper into a more technical discussion of current neurosymbolic systems and methods with their pros and cons. In Section 5, we identify promising approaches and directions for neurosymbolic AI from the perspective of learning, reasoning and explainable AI. In Section 6, we return to the debate that was so present at AAAI-2020 to conclude the paper and identify exciting challenges for the third wave of AI. 

\section{Neurons and Symbols: Context and Current Debate}

Deep learning researchers and AI companies have achieved groundbreaking results in areas such as computer vision, game playing and natural language processing \cite{Hinton-nature,Silver_2017}
Despite the impressive results, deep learning has been criticised for brittleness (being susceptible to adversarial attacks), lack of explainability (not having a formally defined computational semantics or even intuitive explanation, leading to questions around the trustworthiness of AI systems), and lack of parsimony (requiring far too much data, computational power at training time or unacceptable levels of energy consumption) \cite{marcus2020}.
Against this backdrop, leading entrepreneurs and scientists such as Bill Gates and the late Stephen Hawking have voiced concerns about AI's accountability, impact on humanity and the future of the planet \cite{nature-ethics}.
The need for a better understanding of the underlying principles of AI has become generally accepted. A key question however is that of \textbf{identifying the necessary and sufficient building blocks of AI}, and how systems that evolve automatically based on machine learning can be developed and analysed in effective ways that make AI trustworthy. 

Turing award winner and machine learning theory pioneer Leslie Valiant pointed out that a key challenge for Computer Science is the principled combination of reasoning and learning, building a rich semantics and robust representation language for intelligent cognitive behaviour \cite{Valiant2003}. In Valiant's words: 
\emph{``The aim is to identify a way of looking at and manipulating commonsense knowledge that is consistent  with  and  can  support  what  we  consider  to  be  the two most fundamental aspects of intelligent cognitive behaviour: the ability to learn
from experience and the ability to reason from what has been learned. We are therefore seeking
a semantics of knowledge that can computationally support the
basic phenomena of intelligent behaviour."} 
Neural-symbolic computing seeks to offer such a principled way of studying AI by establishing provable correspondences between neural models and logical representations \cite{Bader_2005,Ivan_2017,garcez_book,Garcez_2008,garcez2015}. In neural-symbolic computation, logic can be seen as a language with which to compile a neural network, as discussed in more detail later in this paper.\footnote{Over the years, the terminology ``neural-symbolic'' (integration, computing, system, etc.) was used predominantly by the research community to indicate a combination of two paradigms: neural and symbolic AI, see e.g. \cite{Garcez_2008}
More recently, the more colloquial terminology ``neuro-symbolic" (AI, approach, system, etc) has become more commonly used in publications and the printed press. In this paper, we use the term ``neural-symbolic'' when referring to the combination of paradigms, and we introduce the term ``neurosymbolic" as a single word to symbolise the coming of age of a new area of research.
}

The success of deep learning along with a number of drawbacks identified more recently such as a surprising lack of robustness \cite{intrigue} has prompted a heated debate around the value of symbolic AI by contrast with neural computation and deep learning.
A key weakness, as Bengio et al. state in a recent article, is that \emph{current machine learning methods seem weak when they are required to generalize beyond the training distribution, which is what is often needed in practice} \cite{meta-transfer}.
In the recent AI debate between Yoshua Bengio and Gary Marcus, Marcus argues the case for hybrid systems \cite{marcus2020} and seeks to define what makes an AI system effectively hybrid: 

\begin{quote}
    ``Many more drastic approaches might be pursued. Yoshua Bengio, for example, has made a number of sophisticated suggestions for significantly broadening the toolkit of deep learning, including developing techniques for statistically extracting causal relationships through a sensitivity to distributional changes and techniques for automatically extracting modular structure, both of which I am quite sympathetic to. But for reasons that will become apparent, I worry that even these sorts of tools will not suffice on their own for getting us to robust intelligence.
Instead, I will propose that in order to get to robust artificial intelligence, we need to develop a framework for building systems that can routinely acquire, represent, and manipulate abstract knowledge, with a focus on building systems that use that knowledge in the service of building, updating, and reasoning over complex, internal models of the external world."
\end{quote} 

Key to the appreciation of the above statement by Marcus, also advocated in \cite{DBLP:journals/corr/LakeUTG16} and \cite{GARNELO201917}, is an understanding of the representational value of the symbolic manipulation of variables in logic and the compositionality of language.
It is probably fair to assume that the next decade will be devoted to researching specific methods and techniques which seek to address the above issues of representation, robustness and extrapolation. Such techniques will be drawn from a broader perspective of neurosymbolic machine learning and AI which embraces hybrid systems, including: 

\textbf{(a) Variable Grounding and Symbol Manipulation:} Embracing hybrid systems requires the study of how symbols may emerge and become useful in the context of what deep learning researchers have termed \emph{disentanglement}. Once symbols emerge (which may happen at different levels of abstraction, ideally within a modular network architecture), it may be more productive from a computational perspective to refer to such symbols and manipulate (i.e. compute) them symbolically rather than numerically. Once it becomes known that a complex neural network serves to calculate, for example, the sum of two handwritten digits provided as input images, or equally that a complex neural network has learned the function $f(x)=x$, then it is probably the case that one would prefer such a calculation to be precise and to extrapolate well to any value of $x$. This is easily achieved symbolically. Reasoning, in many cases too, is preferred to be precise and not approximate, although there are cases where approximate or human-like reasoning become more efficient than logical deduction \cite{HalpernUncertainty}. 

\textbf{(b) Commonsense and Combinatorial Reasoning:}
Another key distinction that is worth making explicit refers to the difference between commonsense knowledge and expert knowledge. While the former is approximate and difficult to specify, the latter strives to be as precise as possible and to prove its properties. We believe that, once equipped with a solid understanding of the value of hybrid systems, variable manipulation and reasoning, the debate will be allowed to progress from the question of \emph{symbols versus neurons} to the research question:

\begin{quote}
How to compute and learn with symbols, inside or outside of a neural network, and how efficiently computationally, in a precise or approximate reasoning setting?
\end{quote}

Foundational work about neurosymbolic models and systems such as \cite{garcez_book,NIPS03,Garcez_2008} will be relevant as we embark in this journey. In \cite{Garcez_2008}, correspondences are shown between various logical-symbolic systems and neural network models. The current limits of neural networks as essentially a propositional\footnote{The current limitation of neural networks, which John McCarthy referred to as \emph{propositional fixation}, is of course based on the current simple models of neuron. Although this may be about to change through important work on understanding the mind and brain which may produce richer models of neural networks \cite{Gidon83}, one should note that the recent state-of-the-art results obtained by deep networks using large amounts of data are predicated on the notion of a simple neuron \cite{Hinton-nature,Schmidhuber15}.
}
system are also evaluated. In a nutshell, current neural networks are capable of representing propositional logic, nonmonotonic logic programming, propositional modal logic and fragments of first-order logic, but not full first-order or higher-order logic. This limitation has prompted the recent work in the area of Logic Tensor Networks (LTN) \cite{LTN,marra2019lyrics,vanKrieken2020Analyzing} which, in order to use the language of full first-order logic with deep learning, translates logical statements into the loss function rather than into the network architecture. First-order logic statements are therefore mapped onto differentiable real-valued constraints using a many-valued logic interpretation in the interval [0,1]. The trained network and the logic become communicating modules of a hybrid system, instead of the logic computation being implemented by the network. This distinction between having neural and symbolic modules that communicate in various ways and having translations from one representation to the other in a more integrative approach to reasoning and learning should be at the centre of the debate in the next decade.\footnote{Having worked for two decades on integrative neurosymbolic AI and more recently on hybrid neural-symbolic systems, we are acutely \emph{aware of the tension between principled integration and practical value and application}. Scientifically, there is obvious value in the study of the limits of integration to improve our understanding of the power of neural networks using the well-studied structures and algebras of computer science logic. When seeking to solve a specific problem, however, one may prefer to take, for example, an existing knowledge-base and find the most effective way of using it alongside the tools available from deep learning and software agents. As a case in point, take the unification algorithm, which is an efficient way of computing symbolic substitutions. It is notoriously difficult to implement in neural networks. One may, of course, wish to study how to perform logical unification exactly or approximately using a neural network, although at present the most practical way may be to adopt a hybrid approach whereby unification is computed symbolically.}    

Among the recent neurosymbolic systems, one can identify quite a variety in range from integrative to hybrid systems: \cite{Mao_2019} can be seen as a loosely-coupled hybrid approach where image classification is combined with reasoning from text data; \cite{Robin_2018} offers further integration by allowing a node in the probabilistic inference tree of a symbolic ML system (ProbLog) to be replaced by a neural network; \cite{LTN} takes another step towards integration by using a differentiable many-valued logic in the loss function of a neural network (in LTN, theorem proving is left for the symbolic counterpart of the system); \cite{Minervini} proposes to perform differentiable unification and theorem proving inside the neural network.   

Out of the systems and techniques now available, some more integrative others more loosely-coupled, a common question clearly emerges: \textbf{what are the fundamental building blocks, the necessary and sufficient components of neurosymbolic AI?} For example, is the use of an attention layer necessary \cite{NIPS2017_7181} or can it be replaced by richer structure such as graph networks \cite{lamb2020graph}? Is the explicit use of probability theory necessary, and in this case inside the network or at the symbolic level or both? Is there a real computational gain in combinatorial problem solving by theorem proving using neural networks or is this task better left to the devices of a symbolic system? One thing is now very clearer: there is great practical value in the use of gradient-based learning on distributed representations\cite{10.1145/3065386}.

In this paper, we also seek to bring attention to another perhaps less attractive but equally if not more relevant question of adopting a  \emph{distributed versus a localist representation}. In a localist representation the relevant concepts have an associated identifier. This is typically a discrete representation. By contrast, in a distributed representation, concepts are denoted by vectors with continuous values. This is therefore an issue of which representation is adequate or most appropriate. Symbolic machine learning takes a localist approach while neural networks are distributed, although neural networks can also be localist \cite{localist}. The next section will be devoted to the pros and cons of distributed and localist representations.

\textbf{Forms of Neurosymbolic Integration:} Within neurosymbolic AI one may identify \emph{systems that translate and encode symbolic knowledge in the set of weights of a network} \cite{Franca_2014}, or \emph{systems that translate and encode symbolic knowledge into the loss function of the network}\cite{LTN}. The \emph{neural-symbolic cycle} translating symbolic knowledge into neural networks and vice-versa offers a kind of compiler for neural networks\footnote{In the study of programming languages it is accepted that different levels of abstraction and different representations are needed - e.g. java bytecode and a java program - for the purpose of efficiency, system maintenance, user interaction and verification. We argue that in AI, neural-symbolic systems will provide equally important forms of abstract representation. 
}, whereby prior knowledge is translated into the network, and a decompiler whenever symbolic descriptions are extracted from a trained network. The compiler can either set-up the network's initial weights akin to a one-shot learning algorithm which is guided by knowledge, or define a knowledge-based penalty or constraint which is added to the network's loss function.
 
A third form of integration has been proposed in \cite{meta-transfer} which is based on changing the representation of neural networks into factor graphs. The value of this particular representation deserves to be studied. Change of representation is a worthwhile endeavour on its own right in that it may help us understand the strengths and limitations of different neural models and network architecture choices. This third form of integration, however, proposes to create an intermediate representation with factor graphs in between neural networks and logical representations.  

\textbf{A note about terminology:} In \cite{Pearl}, Turing award winner Judea Pearl offers a critique of machine learning which, unfortunately, conflates the terms \emph{machine learning} and \emph{deep learning}. Similarly, when Geoffrey Hinton refers to \emph{symbolic AI}, the connotation of the term tends to be that of expert systems dispossessed of any ability to learn. The use of the terminology is in need of clarification. Machine learning is not confined to association rule mining, c.f. the body of work on symbolic ML and relational learning \cite{Muggleton_1991} (the differences to deep learning being the choice of representation, localist logical rather than distributed, and the non-use of gradient-based learning algorithms). Equally, symbolic AI is not just about production rules written by hand. A proper definition of AI concerns knowledge representation and reasoning, autonomous multi-agent systems, planning and argumentation, as well as learning. In what follows, we elaborate on the above misunderstandings one at a turn.

\textbf{Symbolic Machine Learning and Deep Learning:} In \cite{Pearl}, Pearl proposes a hierarchy consisting of three levels: association, intervention and counterfactual reasoning, and claims that ML is only capable of achieving association. A neurosymbolic or purely symbolic ML system should be capable of satisfying the requirements of all three of Pearl's levels, e.g. by mapping the neural networks onto symbolic descriptions. It is fair to say in relation to Pearl's top level in the hierarchy - counterfactual reasoning - that progress has only been made recently and that much research is still needed, although good progress is being made towards the extraction of local, measurable counterfactual explanations from black box ML systems \cite{white2019measurable}. Once a neural network has been endowed with a symbolic interpretation, one has no reason to doubt the ability of a neural system to ask \emph{what if} questions. In fact, the very algorithm for extracting symbolic logic descriptions of the form $A \rightarrow B$ from trained neural networks \cite{DBLP:journals/ai/GarcezBG01} uses a form of interrogation of the network akin to the intervention of Bayesian models advocated by Pearl. We argue therefore that a more important question is representational: which representation is most effective, deep networks or Bayesian networks? While attempting to answer this question, as well as considering the demands of the practical applications, it is important to recognise that neural networks offer a concrete model of computation, one which can be implemented efficiently by message passing or propagation of activation, differently from Bayesian networks, and trained by differentiable learning algorithms. A limitation of having such a concrete computational model, however, may be a difficulty of pure neural networks at modelling rich forms of abstraction which are not dependent on the data (images, audio, etc.) but which exist instead at a higher conceptual level. We shall return to this challenge later in the paper.

\textbf{Knowledge Representation and Reasoning in AI:} Complex problem solving using AI requires a much richer language than that of expert systems as suggested by Hinton \cite{hinton}. AI requires a language that can go well beyond Horn clauses to include relational knowledge, time and other modalities, negation by failure, variable substitution and quantification, etc. In statistical relational learning, the use of first-order logic does not require instantiating (or grounding) all possible combinations of the values of the variables (e.g. $X$ and $Y$ in a relation $R(X,Y)$). In \emph{relational reasoning} with neural networks, borrowing from the field of relational databases, it is typically the grounded (and therefore propositional rather than first-order) representation that is learned and reasoned about. For the avoidance of confusion, we would term this latter task \emph{relationship learning}. Two other equally important attributes of a rich language for complex problem solving are \emph{compositionality}, in the sense of the compositionality of the semantics of a logical language, and modularity. It is worth noting that in the original paper about deep learning \cite{Hinton_2006}, before much of the attention turned to convolutional networks, modularity was a main objective of the proposed semi-supervised greedy learning of stacks of restricted Boltzmann machines. The recently-proposed stacked capsule autoencoders \cite{Hinton2019} and neural-symbolic approaches such as Logic Tensor Networks \cite{LTN} as well as other weakly-supervised approaches revive the important stance of modularity in neural computation. Earlier efforts towards modularity in neurosymbolic AI can be traced back to the system for Connectionist Modal and Intuitionistic Logics \cite{GarcezLG06,CML-TCS}. Modal logics with a possible-world semantics have been shown to offer a natural approach to modularity in neural computation \cite{Garcez_2008}.

With AI understood as a superset of ML which in turn is a superset of DL, we shall argue for the combination of statistical machine learning, knowledge representation (KR) and logical reasoning. By logical reasoning, we shall mean not only classical logic reasoning with the traditional true-false interpretation, but non-classical reasoning including nonmonotonic, modal and many-valued logics. In the study of the interplay between learning and reasoning and how best to implement it (e.g. in a continuous or discrete system), it shall become clear that universal quantification is easy to reason about and hard to learn using neural networks; existential quantification is easy to learn and harder to reason about in a symbolic system. Such limitations on either side of the spectrum will dictate a few practical design decisions to be discussed in this paper. In a nutshell, \emph{we claim that neurosymbolic AI is well placed to address concerns of computational efficiency, modularity, KR + ML and even causal inference}. More researchers than ever on both sides of the connectionist-symbolic AI divide are now open to studying and learning about each others' tools and techniques. This was not the case until very recently. The use of different terminology alongside some preconceived opinion or perhaps idleness, fuelled by the way that science normally rewards research that is carried out in silos, have prevented earlier progress. The fact that this is now changing will lead to faster progress in the overall field of AI. It is reassuring to see it happening in this way: the neural information processing community have shown the value of neural computation in practice, which has attracted the curiosity of great minds from symbolic AI. We hope that further collaboration in neurosymbolic AI will help solve many of the issues which are still outstanding.

\section{Distributed and Localist Representation}

The integration of learning and reasoning through neurosymbolic systems requires a bridge between localist and distributed representations. The success of deep learning indicates that distributed representations with gradient-based methods are more adequate than localist ones for learning and optimization. At the same time, the difficulty of neural networks at extrapolation, explainability and goal-directed reasoning point to the need of a bridge between distributed and localist representations for reasoning. 

Neural-symbolic computing has been an active area of research seeking to establish such a bridge for several years \cite{Bader_2005,Evans_18,garcez_book,Garcez_2008,hammer-hitzler,Khardon_97,Serafini_2016,Valiant_2006}. In neural-symbolic computation, knowledge learned by a neural network can be represented symbolically. Reasoning takes place either symbolically or within the network in distributed form. Despite their differences, both the symbolic and connectionist paradigms share common characteristics, offering benefits when put together in a principled way (see e.g. \cite{NIPS03,Garcez_2008,Smolensky_1995,Valiant_2006}). Change of representation also offers a way of making sense of the value of different neural models and architectures with respect to what is a more formal and better understood area of research: symbolic logic.

Neural network-based learning and inference under uncertainty have been expected to address the brittleness and computational complexity of symbolic systems. Symbolism has been expected to provide additional knowledge in the form of constraints for learning \cite{Ivan_2017,Gori}, which ameliorate neural network's well-known catastrophic forgetting or difficulty with extrapolation in unbounded domains or with out-of-distribution data. The integration of neural models with logic-based symbolism is expected therefore to provide an AI system capable of explainability, transfer learning and a bridge between lower-level information processing (for efficient perception and pattern recognition) and higher-level abstract knowledge (for reasoning, extrapolation and planning).

Suppose that a complex neural network learns a function $f(x)$. Once this function is known, or more precisely a simplified description of $f(x)$ is known, computationally it makes sense to use such a representation, not least for the sake of extrapolation, as exemplified earlier with the $f(x)=x$ function. One could argue that at this point the neural network has become superfluous. Symbol manipulation (once symbols have been discovered) is key to further learning at new levels of abstraction. This is exemplified well in \cite{marcus2020} with the use of the concept of a \emph{container} which may be learned from images.   

Among the most promising recent approaches to neural-symbolic integration, so-called embedding techniques seek to transform symbolic representations into vector spaces where
reasoning can take place through matrix computations over distance functions \cite{Bordes_2011,Socher_2013,Ilya_2008,Serafini_2016,Santoro_2017,Cohen_2017,Evans_18,Yang_2017,Dong_2019,Rocktaschel_2016}. In such systems, learning of an embedding is carried out using backpropagation \cite{Werbos:74,Rumelhart:1986we}. Most of the research in this area is focused on the art of representing relational knowledge such as $P(X,Y)$ in a distributed neural network. The logical predicate $P$ relating variables $X$ and $Y$ could be used to denote, for example, the \emph{container} relation between two objects in an image such as a violin and its case, which are in turn described by their embedding. This process is known as relational embedding \cite{Bordes_2011,Santoro_2017,Socher_2013,Ilya_2008}. For representing more complex logical structures such as first order-logic formulas, e.g. $\forall X,Y,Z: (P(X,Y) \rightarrow Q(Y,Z))$, a
system named {\it Logic Tensor Networks} (LTN) \cite{Serafini_2016} was proposed by extending {\it Neural Tensor Networks} (NTN)
\cite{Socher_2013}, a state-of-the-art relational embedding
method. Related ideas are discussed formally in the context of 
constraint-based learning and reasoning in \cite{Gori}.

Two powerful concepts of LTN are (1) the grounding of logical concepts onto tensors with the use of logical statements which act as constraints on the vector space to help learning of an adequate embedding, and (2) the modular and differentiable organisation of knowledge within the neural network which allows querying and interaction with the system. Any user-defined statement in first-order logic can be queried in LTN which checks if that knowledge is satisfied by the trained neural network. With such a tool, a user can decide when to keep using a distributed connectionist representation or switch to a localist symbolic representation. This last aspect brings the question of the emergence of symbols and their meaning in neural networks to the fore: recent work using the weakly supervision of auto-encoders and ideas borrowed from disentanglement have been showing promise in the direction of learning relevant concepts which can in turn be re-used symbolically \cite{chen2019weakly}. Related work seeking to explore the advantages of distributed representations of logic include \cite{Cohen_2017}, which is based on stochastic logic programs, \cite{Evans_18,Yang_2017,Dong_2019}, with a focus on inductive programming, and \cite{Rocktaschel_2016}, based on differentiable theorem proving.

\textbf{A taxonomy for neurosymbolic AI:} with an understanding of the role of localist and distributed approaches, we now provide an analysis of Henry Kautz's taxonomy for neurosymbolic AI \cite{HenryKautz}, which was introduced at AAAI 2020: In Kautz's taxonomy, a \textsc{Type 1} neural-symbolic integration is standard deep learning, which some may argue is a stretch, but which is included by Kautz to note that the input and output of a neural network can be made of symbols, e.g. text in the case of language translation or question answering applications. \textsc{Type 2} are hybrid systems such as DeepMind's AlphaGo and other systems where the core neural network is loosely-coupled with a symbolic problem solver such as Monte Carlo tree search. \textsc{Type 3} is also a hybrid system whereby a neural network focusing on one task (e.g. object detection) interacts via its input and output with a symbolic system specialising in a complementary task (e.g. query answering). Examples include the neuro-symbolic concept learner \cite{Mao_2019} and deepProbLog \cite{Robin_2018}. In a \textsc{Type 4} neural-symbolic system, symbolic knowledge is compiled into the training set of a neural network. Kautz offers \cite{Lample2020Deep} as an example (to be read alongside the critique in \cite{ernie}). An approach to learn and reason over mathematical constructions is proposed in \cite{ArabshahiSA18}, and in \cite{Arabshahi19} a learning architecture that extrapolates to harder symbolic maths reasoning problems is introduced.
We would also include in \textsc{Type 4} other tightly-coupled but localist neural-symbolic systems where various forms of symbolic knowledge, not restricted to \emph{if-then} rules, is translated into the initial architecture and set of weights of a neural network, in some cases with guarantees of correctness \cite{Garcez_2008}, as well as Logical Neural Networks, where the key concept is to create a 1-to-1 correspondence between neurons and the elements of logical formulas \cite{Fagin2020}.
\textsc{Type 5} are those tightly-coupled but distributed neural-symbolic systems where a symbolic logic rule is mapped onto an embedding which acts as a soft-constraint (a regularizer) on the network's loss function. Examples of these include Logic Tensor Networks \cite{LTN} and Tensor Product Representations \cite{Smolensky}, referred to in \cite{JAL19} as \emph{tensorization} methods. Finally, a \textsc{Type 6} system should be capable, according to Kautz, of \emph{true symbolic reasoning inside a neural engine}. This is what one could refer to as a fully-integrated system. Early work in neural-symbolic computing has achieved this (see \cite{Garcez_2008} for a historical overview). Some \textsc{Type 4} systems are also capable of it, but using a localist rather than a distributed representation and using much simpler forms of embedding than \textsc{Type 5} systems. Kautz adds that a \textsc{Type 6} system should be capable of \emph{combinatorial reasoning}, possibly by using an attention schema to achieve it effectively. Recent efforts in this direction include \cite{leytonAAAI2020,lamb2020graph,prates2019tsp}, although a fully-fledged \textsc{Type 6} system for combinatorial reasoning does not exist yet.

Further research into \textsc{Type 5} systems will likely focus on the provision of rich embeddings and the study of the extent to which such embeddings may correspond either to pre-defined prior knowledge or to learned attention mechanisms. Further research onto \textsc{Type 6} systems is highly relevant to the theory of neural-symbolic computing, as discussed in more detail in the next section. In practical terms, a tension exists between effective learning and sound reasoning, which may prescribe the use of a more hybrid approach of \textsc{Types 3 to 5}, or variations thereof such as the use of attention with tensorization. Orthogonal to the above taxonomy, but mostly associated thus far with \textsc{Type 4}, is the study of the limits of reasoning within neural networks, which was already of interest since the first efforts by Valiant at providing a foundation for computational learning \cite{Valiant_1984}. Recently, this has been the focus of experimental analyses of deep learning in symbolic domains \cite{Tavares2020}, and it should include the study of first-order logic, higher-order, many-valued and non-classical logic. 

\section{Neurosymbolic Computing Systems: Technical Aspects}
In symbolic ML, symbols are manipulated as part of a discrete search for the best representation to solve a given classification or regression task. The most well-known form of symbolic ML are decision trees, but richer forms of representation exist, in particular relational representations using first-order logic to denote concepts ranging over variables $X,Y,Z...$ within a (possibly infinite) domain, e.g. $\forall X,Y,Z: grandfather(X,Y) \leftarrow (father(X,Z) \wedge mother(Z,Y))$ (the father of someone's mother is that person's grandfather). Probabilistic extensions of this approach seek to learn probability distributions for such logical rules (or functional programs) as a way of accounting for uncertainty in the training data. Work in these areas is probably best characterised by the conference series on Inductive Logic Programming \cite{DBLP:series/synthesis/2016Raedt,DBLP:conf/ijcai/MuggletonL13}, Statistical Relational Learning  \cite{DBLP:journals/ml/RichardsonD06,10.5555/3208509,DBLP:journals/corr/BachBHG15, shao2019conditional} and Probabilistic or Inductive Programming \cite{DBLP:journals/dagstuhl-reports/SchmidMS17}. 

All of the excitement and industrial interest in the past 10 years surrounding AI and Machine Learning, though, have come from an entirely separate type of ML: deep learning. Deep learning uses neural networks and stochastic gradient descent to search through a continuous space, also to solve a given classification or regression task, but creating vector-based, distributed representations, rather than logical or symbolic ones. For this reason, such systems are called sub-symbolic.

Whilst it is clear now that AI will not be achieved by building expert systems by hand from scratch (GOFAI), but by learning from large collections of data, one would be misguided to conflate all of machine learning or dismiss the role of symbolic logic, which remains the most powerful and adequate representation for the analysis of computational systems. As put simply by Moshe Vardi \emph{``Logic is the Calculus of Computer Science''} and, differently from statistics, machine learning can only exist within the context of a computational system. 

Specifically, deep neural networks will require a language for description, as also advocated by Leslie Valiant. \emph{Neural network-based AI is distributed and continuous}, deals well with large-scale multimodal noisy perceptual data such as text and audio, handles symbol grounding better than symbolic systems since concepts are grounded on feature vectors, and is by definition a computational model, frequently implemented efficiently using propagation of activation and tensor processing units.\footnote{Contrast with Bayesian networks which may be inefficient as a computational model frequently requiring simplification of graphs into tree-based representations.} \emph{Symbolic AI is generally localist and discrete, capable of sophisticated reasoning}, including temporal, epistemic and nonmonotonic reasoning, planning, extrapolation and reasoning by analogy. Neurosymbolic AI has shown that non-classical logics, in particular many-valued logics, offer an adequate language for describing neural networks \cite{LTN, Fagin2020}. As the field of AI moves towards agreement on the need for combining the strengths of neural and symbolic AI, it should turn next to the question: \textbf{what is the best representation for neurosymbolic AI?}

To answer this question, one should seek to be informed by developments in neural-symbolic computing of the past 20 years, and to evaluate in a precise manner the methods, algorithms and applications of neurosymbolic AI. For instance, it is known that current recurrent neural networks are capable of computing the logical consequences of propositional modal logic programs and other forms of non-classical reasoning and fragment of first order logic programs \cite{DBLP:series/sci/BaderHHW07, Garcez_2008}. Obtaining results for full first-order logic has not been possible thus far, which reinforces John McCarthy's claim that neural networks are essentially propositional. In terms of applications of AI, these have been largely focused on perceptual or pattern matching tasks such as image and audio classification. Recent efforts at question answering and language translation as well as protein folding classification have highlighted the importance of the neurosymbolic approach. The ideal type of application for a neurosymbolic system, however, should be that where abstract information is required to be reasoned about at different levels beyond that what can be perceived from data alone, such as complex concept learning whereby simpler concepts are required to be organised systematically as part of the definition of a higher concept. Such a conceptual structure, which is still to be discovered using data, also requires knowledge which is governed by \emph{general rules and exceptions to the rules}, allowing for sound generalization in the face of uncertainty but also capable of handling specific cases (the many exceptions, which may be important for the sake of robustness although not necessarily statistically relevant). Similarly, it seems hard to achieve true relational learning using only neural networks. A useful but simple example can be borrowed from the area of Inductive Logic Programming: learning the concept of \emph{ancestor} from a few examples of the \emph{mother}, \emph{father} and \emph{grandparent} relations. Grounding the entire knowledge-base in this case would not be productive since the chain of reasoning to derive the concept of \emph{ancestor} may be arbitrarily large depending on the data available. In this case, one is better off learning certain relations by \emph{jumping to conclusions}, such as e.g. $\forall X,Y: father(X,Y) \rightarrow ancestor(X,Y)$, from relatively few examples and using similarity measures to infer new relations, at the same time deriving symbolic descriptions which can be used for reasoning beyond the distribution of the data, allowing in turn for extrapolation. In this example, once a description for \emph{ancestor} is obtained, one should be able to reason about arbitrarily long chains of family relationships. Notice that key to this process is the ability to revise the conclusion taken once new evidence to the contrary of what has been inferred is made available from the data. In other words, the reasoning here is nonmonotonic \cite{Garcez_1999}. 

In summary, \textbf{at least two options exist for neurosymbolic AI}. In Option 1, symbols are translated into a neural network and one seeks to perform reasoning within the network. In Option 2, a more hybrid approach is taken whereby the network interacts with a symbolic system for reasoning. A third option, which would not require a neurosymbolic approach, exists when expert knowledge is made available, rather than learned from data, and one is interested in achieving precise sound reasoning as opposed to approximate reasoning. We discuss each option briefly next.

In Option 1, it is desirable still to produce a symbolic description of the network for the sake of improving explainability (discussed later) or trust, or for the purpose of communication and interaction with the system. In Option 2, by definition, a neurosymbolic interface is needed. This may be the best option in practice given the need for combining reasoning and learning in AI, and the apparent different nature of both tasks (discrete and exact versus continuous and approximate). However, the value of distributed approximate reasoning using neural networks is only starting to be investigated as in the case of differentiable neural computers \cite{DBLP:journals/nature/GravesWRHDGCGRA16} and neural theorem proving \cite{Minervini2}, although early efforts did not prove to be promising in terms of practical efficiency \cite{Pinkas_1995,Hoho90,Shastri}. In Option 3, a reasonable requirement nowadays would be to compare results with deep learning and the other options. This is warranted by the latest practical results of deep learning showing that neural networks can offer, at least from a computational perspective, better results than purely symbolic systems. 

In practice, the choice between Options 1 and 2 above may depend on the application at hand and the availability of quality data and knowledge. A comparatively small number of scientists will continue to seek to make sense of the strengths and limitations of both neural and symbolic approaches. On this front, the research advances faster on the symbolic side due to the clear hierarchy of semantics and language expressiveness and rigour that exists at the foundation of the area. By contrast, little is known about the expressiveness of the latest deep learning models in relation to established neural models beyond data-driven comparative empirical evaluations. As advocated by Paul Smolensky, neurosymbolic computing can help map the latest neural models into existing symbolic hierarchies, thus helping organise the extensively ad-hoc body of work in neural computation. 

\section{Challenges for the Principled Combination of Reasoning and Learning} 
For a combined perspective on reasoning and learning, it is useful to note that reasoning systems may have difficulties computationally when reasoning with existential quantifiers and function symbols, such as $ \exists x P(f(x))$. Efficient logic-based programming languages such as Prolog, for example, assume that every logical statement is universally quantified. By contrast, learning systems may have difficulty when adopting universal quantification over variables. To be able to learn a universally quantified statement such as $ \forall x P(x)$, a learning systems needs in theory to be exposed to all possible instances of $x$. This simple duality points to a possible complementary nature of the strengths of learning and reasoning systems. To learn efficiently $ \forall x P(x)$, a learning system needs to \emph{jump to conclusions}, extrapolating $\forall x P(x)$ given an adequate amount of evidence (the number of examples or instances of $x$). Such conclusions may obviously need to be revised over time in the presence of new evidence, as in the case of nonmonotonic logic. In this case, a statement of the form $\forall x P(x)$ becomes a data-dependent generalization, which is not to be assumed equivalent to a statement $\forall y P(y)$, as done in classical logic. Such statements may have been learned from different samples of the overall potentially infinite population. On the other hand, a statement of the form $\exists x P(x)$ is trivial to learn from data by identifying at least one case $P(a)$, although reasoning from $\exists x P(x)$ is more involved, requiring the adoption of an arbitrary constant $b$ such that $P(b)$ holds.

It is now accepted that learning takes place on a continuous search space of (sub)differentiable functions; reasoning takes place in general on a discrete space as in the case of goal-directed theorem proving. The most immediate way of benefiting from the combination of reasoning and learning, therefore, is to adopt a hybrid approach whereby a neural network takes care of the continuous search space and learning of probabilities, while a symbolic system consisting of logical descriptions of the network uses discrete search to achieve extrapolation and goal-directed reasoning.\footnote{The investigation of continuous reasoning or discrete learning approaches is of course worth pursuing too. There have been a number of recent developments with relevant results in both of these directions \cite{DBLP:journals/corr/RocktaschelR17,Robin_2018,Yang_2017,raedt2020statistical}. Computational complexity issues remain a challenge though for the combination of first-order logic and probabilities in symbolic AI.} As hinted already in this paper, a property of early deep learning may hold the key to the above hybrid perspective: modularity. In the original paper on deep learning \cite{Hinton_2006}, a modular system is proposed consisting of a stack of restricted Boltzmann machines (RBMs). The extraction of symbolic descriptions from each RBM is thus made considerably easier \cite{Son_2018}. Each RBM learns a joint probability distribution while their symbolic description reflects the result of learning without manipulating probabilities explicitly, thus avoiding the complexity of probabilistic inference found in symbolic AI.

In \cite{Son_2018}, an efficient algorithm is presented that extracts propositional rules enriched with confidence values from RBMs, similar to what was proposed with Penalty Logic for Hopfield networks in \cite{Pinkas_1995}. When RBMs are stacked onto a deep belief network, however, the modular extraction of compositional rules may be accompanied by a compounding loss of accuracy, indicating that knowledge learned by the neural network might not have been as modular as one would have wished. Systems that impose a more explicit separation of modules may hold the answer to this problem, in particular systems where unsupervised learning is combined with weakly-supervised classification at distinct levels of abstraction, such as with the use of variational auto-encoders \cite{stroemfelt2020transferability} or generative adversarial networks \cite{NIPS2014_5ca3e9b1}. In the area of Reinforcement Learning too, deep learning systems combined with symbolic rules were first proposed in \cite{DBLP:journals/corr/GarneloAS16}, and there is promise of considerable progress with the use of neurosymbolic approaches especially in the case of model-based Reinforcement Learning.     

We therefore do not advocate the adoption of \emph{monoblock} networks with millions of parameters. Even though this may be how the human brain works, loss of modularity seems to be, at least at present from a computational perspective, a price that is too high to pay. Modularity remains a fundamentally relevant property of any computing system.

\textbf{Applications of Neurosymbolic AI:} One major way of driving advances in AI continues to be through challenging applications, be it language translation, computer games or protein folding competitions. Language understanding in the broadest sense of the term, including question-answering that requires commonsense reasoning, offers probably the most complete application area for neurosymbolic AI. As an example, consider this question and its commonsense answer from the COPA data set \cite{roemmele_choice_2011}: It got dark outside. What happened as a result? (a) Snowflakes began to fall from the sky; (b) The moon became visible in the sky. Another very relevant application domain is planning, which requires learning and reasoning over time, as in this example adapted from \cite{DBLP:journals/corr/abs-1811-12143}: Daniel picks up the milk; Daniel goes to the bedroom; Daniel places the milk on the table; Daniel goes to the bathroom. Where is the milk? Finally, an area where 
machine learning and knowledge representation and reasoning have complementary strengths is \emph{knowledge engineering}, including knowledge-base completion and data-driven ontology learning. In this area of application, rich and large-scale symbolic representations exist alongside data, including knowledge graphs to be combined with neural networks such as graph neural networks. 

A common thread across the above examples and applications is the need for modelling \emph{cause and effect} with the use of implicit information. This requires learning of general rules and exceptions to the rules that evolve over time. In such cases, deep learning alone fails when presented with examples from outside the distribution of the training data. This motivated Judea Pearl's critique of Machine Learning \cite{Pearl} which we shall address in some detail next.

In Pearl's 3-level causal hierarchy (association, intervention and counterfactuals), association involves purely statistical relationships. In Pearl's words, \emph{observing a customer who buys toothpaste makes it more likely that this customer will also buy floss. Such associations can be inferred directly from the observed data using standard conditional probabilities and conditional expectation or other standard non-probabilistic ML model. Questions asked at this level require no causal information and, for this reason, this layer is placed at the bottom of the hierarchy. Answering such questions is the hallmark of current machine learning methods}. Pearl's hierarchy may unintentionally give the impression that machine learning is confined to this bottom layer, since no reference is made in \cite{Pearl} to the body of work on symbolic machine learning which is unequivocally not confined to association rules \cite{Muggleton_1991,Rocktaschel_2016}. Pearl continues: \emph{the second level, intervention, ranks higher than association because it involves not just seeing what is but changing what we see. A typical question at this level would be: what will happen if we double the price? Such a question cannot be answered from sales data alone, as it involves a change in customers' choices in reaction to the new pricing. These choices may differ substantially from those taken in previous price-raising situations, unless we replicate precisely the market conditions that existed when the price reached double its current value}. At this level there is a need for inference that can reach beyond the data distribution. Finally, the top level invokes counterfactuals, a mode of reasoning that reverts to the philosophers David Hume and John Stuart Mill and that has been given computational semantics in the past two decades. A typical question in the counterfactual category is: \emph{what if I had acted differently?}, thus necessitating retrospective reasoning.

As noted earlier in the paper, neural-symbolic computing can implement all three of Pearl's levels. Once a symbolic description of the form \emph{if A then B} has been associated with a neural network, surely the idea of intervention \cite{DBLP:conf/kdd/CaruanaLGKSE15} and counterfactual reasoning become possible, c.f. for example, \cite{white2019measurable} on the measurement and extraction of counterfactual knowledge from trained neural networks. 

\textbf{Our conclusion from the above discussion is that in neurosymbolic AI}: 

\begin{itemize}
    \item Knowledge should be grounded onto vector representations for efficient learning from data based on message passing in neural networks as an efficient computational model. 
    \item Symbols should become available as a result of querying and knowledge extraction from trained networks, and offer a rich description language at an adequate level of abstraction, enabling \emph{infinite uses of finite means}, but also compositional discrete reasoning at the symbolic level allowing for extrapolation beyond the data distribution.
    \item The combination of learning and reasoning should offer an important alternative to the problem of combinatorial reasoning by learning to reduce the number of effective combinations, thus producing simpler symbolic descriptions as part of the neurosymbolic cycle.  
\end{itemize}

As an example, consider an Autoencoder which learns in unsupervised fashion to maximise mutual information between pixel inputs and a latent code or some other embedding consisting of fewer relevant features than pixels. Suppose that this neural network has learned to find regularities such as e.g. \emph{when it sees features of type A, it also sees features of type B but not features of type C}. At this point, such regularities can be converted into symbols: $\forall x A(x) \rightarrow (\exists y B(y) \wedge \neg \exists z C(z))$. As a result of the use of variables $x,y,z$ at the symbolic level, one can extrapolate the above regularity to any features of type $A$, $B$ or $C$.\footnote{As another example, consider a neural network trained to classify graphs into those which contain a Hamiltonian cycle and those which do not, given a fixed range of available graph sizes as training examples. Contrast this network with a symbolic description of the definition of Hamiltonian cycle which therefore applies to graphs of any size.} A symbolic description is also a constraint on the neurosymbolic cycle. It is generalised from data during learning and it certainly includes an ability to ask what-if questions. Reasoning about what has been learned allows for extrapolation beyond the data distribution, and finally the symbolic description can serve as prior knowledge (as a constraint) for further learning in the presence of more data, which includes the case of knowledge-based transfer learning \cite{Son_2018}. Further training can now take place at the perceptual or sub-symbolic level, or at the conceptual or symbolic level. This is when having a distributed (sub-symbolic) and a localist (symbolic or sub-symbolic) representation becomes relevant. Assuming that probabilities are dealt with at the sub-symbolic distributed level (as in RBMs) and that the symbolic level is used for a more qualitative representation of uncertainty in the form of general rules with exceptions, we avoid the complications of having to deal simultaneously with discrete and continuous learning of rules and probabilities. 

\textbf {A Note about Explainable AI (XAI):} Knowledge extraction is an integral part of neural-symbolic integration and a major ingredient towards explainability of black-box AI systems. The main difficulty in XAI is the efficient extraction of compact and yet correct and complete knowledge. It can be argued that a large knowledge-base is not more explainable than a large neural network.\footnote{The now apparent lack of explainability of Random Forests, which amount to a collection of Decision Trees and therefore propositional logic formulas with probabilities, serves as a good reminder that XAI is not confined to neural networks.} Although this may be true at the level of the entire model explainability, in the case of local explanations, i.e. explanations of individual cases, a knowledge-base is certainly more explainable than a neural network because it offers a trace (a proof history) showing why an outcome was obtained, as opposed to simply showing how propagation of activation through the network has led to that outcome. It is a main goal of knowledge extraction algorithms to seek to derive compact relevant representations from large complex networks. This is not always possible to achieve efficiently, in which case one may need to resort to having local explanations only.

Many aspects of XAI are being investigated at present. Some of the research questions include: Is the explanation intended for an expert or lay person? Is an explanation required because one does not trust the system, expected a different outcome/would like to induce a different outcome, or would like to question the normative system that has led to the outcome? Is an explanation intended to try and improve system performance, reduce bias/increase fairness, or is it the case that one would simply like to be able to understand the decision process? 
The answers to these questions are likely to be application and user specific. While some stakeholders may be happy without an explanation, as in the case of a patient faced with a medical diagnosis, in most cases some form of explanation is needed to improve system performance or trust. In some cases, producing an explanation is in fact the main goal of the system as in the application of ML to responsible gambling in \cite{DBLP:conf/ecai/PercyGDFSW16,percy}.

Early efforts at knowledge extraction in neurosymbolic AI as a form of explanation were always evaluated w.r.t. fidelity: a measure of the accuracy of the extracted knowledge in relation to the neural network rather than the data, or in today's terminology, a measure of the accuracy of the \emph{student} model w.r.t. the \emph{teacher} model. 
High fidelity is therefore fundamental whenever a student model is to be claimed to offer a good explanation for a (more complex) teacher model. Unfortunately, many recent XAI methods have abandoned fidelity as a measure of the quality of an explanation, making it easier for an apparently excellent explanation to be simply wrong, in that it may not be at all an explanation of the ML model in question. This is particularly problematic for very popular local XAI methods such as LIME \cite{DBLP:conf/kdd/Ribeiro0G16}. Local explanation systems currently in use by consultancy firms and available in many AI toolboxes can be shown to achieve very low levels of fidelity. Without high fidelity results, an apparently perfectly good explanation produced by an XAI system is likely not to be an explanation of the underlying ML system which it is expected to explain. In \cite{white2019measurable} a way of measuring fidelity of local methods was introduced which we argue should be adopted by all XAI methods. The same paper exemplifies how LIME's explanations may achieve very low fidelity. 
 
Even better than fidelity, if an XAI method can be shown to be sound \cite{DBLP:journals/ai/GarcezBG01} then it will provably converge to a high-fidelity. Soundness however is normally 
associated with exponential complexity and so in practice a measure of fidelity may be all that is available. Knowledge extraction should also allow communication between users and the ML system. Current interactive ML is insufficient to the extent that it proposes to replace sound statistical evaluation by subjective user evaluation. Communication with the system implies an ability to ask questions (query the system) and check one's understanding (obtain a rationale for the outcome). The user can then either agree with the outcome and the rationale, agree with the outcome but not the rationale, or disagree with the outcome, thus providing useful feedback or direct intervention to change the system and its outcomes, probably through the system's symbolic description language in the case of neurosymbolic AI. 

Knowledge extraction also offers a way of identifying and correcting for bias in the ML system, which is a serious and present problem \cite{10.5555/3208509}. As a result of the General Data Protection Regulation (GDPR), many companies have decided as a precaution to remove protected variables such as gender and race from their ML system. It is well known, however, that proxies exist in the data which will continue to bias the outcome so that the removal of such variables may serve only to hide a bias that otherwise could have been revealed via knowledge extraction \cite{pessach2020algorithmic}.
Current AI-based decision support systems process very large amounts of data which humans cannot possibly evaluate in a timely fashion. Thus, even with a so-called human-in-the-loop approach where technology domain experts or end-users may become accountable to the decisions, domain experts or end-users are likely to quickly feel progressively less capable of over-riding recommendations which were deemed accurate and based on so much more data than they can handle. Even when such AI systems are portrayed as decision support systems, the current reality is that in order to function well with Big Data, the system must execute a form of triage of the data to be presented to the expert, thus only offering partial information to the decision maker. Without knowledge extraction and a capacity for system communication, the decision maker will, by the very nature of the automated data triage, not be in control.   
Finally, the simple extraction of rules from trained networks may be insufficient. One may need to extract also \emph{confidence values} so as to be able to rank extracted rules. This offers a system that \emph{knows when it does not know}. As a simple example, consider a typical neural network trained to classify the well-known MNIST hand-written digits from 0 to 9. Faced with an image of an obvious non-digit such as an image of a cat, this system must surely provide a very low confidence value to any of the outcomes 0 to 9. The use of adversarial approaches alongside knowledge extraction for robustness has a contribution to make here. In summary, for the many reasons discussed above, neurosymbolic AI with a measurable form of knowledge extraction is a fundamental part of XAI.

\section {AAAI 2020, a Turning Point}
We now return to the debate around System 1 and System 2 that motivated the introduction of this paper and provide a short summary of the AAAI 2020 conference. Not only did the AAAI 2020 conference contained a larger than usual number of papers proposing to combine neural networks and symbolic AI, there were a number of keynote addresses and debates directly relevant to neural-symbolic computing, notably:\\
(1) Fireside conversation between 2019 Turing award winners, Francesca Rossi, and Nobel Laureate Daniel Kahneman on thinking fast and slow and its relation to neural networks and symbol manipulation; \\
(2) The Third AI Summer, Henry Kautz's AAAI 2020 Robert S. Engelmore Memorial Award Lecture, which introduced a taxonomy for neurosymbolic computing;\\ (3) The Director of MIT/IBM Watson AI laboratory David Cox IAAI keynote address, which focused on neurosymbolic AI and applications on vision and language understanding, machine commonsense, question answering, argumentation and XAI.

At the Turing award session and fireside conversation with Daniel Kahneman, there was a clear convergence towards integrating symbolic reasoning and deep learning. Kahneman made his point clear by stating that on top of deep learning, a System 2 symbolic layer is needed. 
This is reassuring for neurosymbolic AI going forward as a single more cohesive research community that can agree about definitions and terminology, rather than a community divided as AI has been up to now. 
 
AI is still in its infancy, so perhaps some of the early disputes can be understood. The debate around \emph{symbols versus neurons} is unlikely to produce concrete results unless it prompts researchers on either side of the divide to learn about each others' methods and techniques. As the saying goes, ``all vectors are symbols, but not all symbols are vectors''. 

Kahneman made the point that System 1 (S1) and System 2 (S2) are terms not coined by him which have a long history in psychology research \cite{Kahneman2011}, and that he prefers to use \emph{implicit} versus \emph{explicit} thinking and reasoning \cite{fireside2020}. 
He argued that S1 (as the intuitive parallel system) is capable of understanding language, in contradiction with Yoshua Bengio's account of deep learning's S1 and S2. Kahneman also stated that S2 (as the sequential deliberative system) is most probably performing symbol manipulation as argued by Gary Marcus in \cite{marcus2020}. In his ``next decade of AI" paper, Marcus argues strongly in favour of hybrid systems, and seeks to define what makes a system hybrid. Marcus's definition is important inasmuch as a main problem of deep learning is a lack of definition. By contrast, Yann LeCun's recent attempt at defining deep learning, also presented at the AAAI 2020 debate \cite{fireside2020}, falls short of what is a useful formal definition. For example, LeCun's definition fails to distinguish deep and shallow networks. 
All attempts to create such a bridge between S1 and S2 are at this point useful and should be commended given our state of lack of understanding of how the brain works. For example, attempts to create differentiable reasoning are useful, although we would require an important distinction be made, whether the purpose is to achieve brain-like systems or to create robust AI. It is possible that these two goals may soon lead into two quite separate research directions: those who seek to understand and model the brain and those who seek to achieve or improve AI. Maybe from that perspective the field is too broad and will require further specialization. A common challenge that will persist, however, is embodied in the question: \textbf{how symbolic meaning emerges from large networks of neurons?} 
 
Perhaps an important choice for neurosymbolic AI is the choice between combinatorial (exact) reasoning and commonsense, approximate reasoning. While learning is always approximate, reasoning can be approximate or precise. In a neurosymbolic system, it is possible to envisage the combination of efficient approximate reasoning (jumping to conclusions) with more deliberative and precise or normative symbolic reasoning \cite{Valiant2013}. Conclusions may be revised through learning from new observations and via communication with the system through knowledge extraction and precise reasoning. One might expect commonsense to emerge as a result of this process of reasoning and learning, although the modelling and computing of commonsense knowledge continues to be another challenge.

From a practical perspective, a recipe for neurosymbolic AI might be: learning is carried out from data by neural networks which use gradient-descent optimization; efficient forms of propositional reasoning can also be carried out by the network, c.f. neural-symbolic cognitive reasoning \cite{Garcez_2008}; rich first-order logic reasoning and extrapolation needs to be done symbolically from descriptions extracted from the trained network; once symbolic meaning has emerged from the trained network, symbols can be manipulated easily by the current computer and can serve as constraints for further learning from data as done in \cite{LTN}; this establishes a practical form of neurosymbolic cycle for learning and reasoning which is feasible with the current technology.

\textbf{Ingredients of neurosymbolic AI:} Narrow AI based on neural networks is already successful and useful in practice with big data. There is obvious value in this as shown by the flourishing of the Machine Learning community and the growing NeurIPS conference community. Data scientists will do this work. In this paper, however, we have been discussing the science of what constitutes the fundamental ingredients of an intelligent system \cite{Valiant2003}. One such ingredient, current results show, is gradient-based optimization used by deep learning to handle large amounts of data, but other ingredients are surely needed. At the AAAI-2020 fireside conversation, a question was asked about the beauty and value of abstract compact symbolic representations such as $F=ma$ or $e=mc^{2}$. Yoshua Bengio's answer was to point out that these must have come out of someone's brain, Isaac Newton and Albert Einstein to be precise. As Stephen Muggleton noted at another debate on the future of AI, his goal is to shorten the wait for the next Newton or Einstein, or Alan Turing. Muggleton's bet is on the use of higher-order logic representations and meta-interpretive learning.\footnote{In a recent paper, Pedro Domingos has shown that gradient descent achieves in effect a superposition of the training examples, similar to a data-dependent form of kernel \cite{domingos2020model}, which in our view highlights the importance of the structure of neural networks, and not just their function.} With this example we seek to illustrate that among highly respected researchers the choice of ingredients may vary widely, from the need for much more realistic models of the brain to the need for ever more sophisticated forms of higher-order computation. With the neural-symbolic methodology, the goal is to develop neural network models with a symbolic interpretation. The key is how to learn representations neurally and make them available for use symbolically (as for example when an AI system is asked to explain itself). In this paper, we have argued for modularity as an important ingredient, allowing one to refer to large parts of the network by the composition of symbols and relations among them. Having an adequate language for describing knowledge encoded in such networks is another important ingredient. We have argued for the use of first-order logic as this language, as a canonical form of representation, but also other forms of non-classical representation such as nonmonotonic and modal logic and logic programming. Once a complex network can be described symbolically, ideally in an abstract compact form as in $e=mc^{2}$, any style of deductive reasoning becomes possible. Reasoning is obviously another fundamental ingredient, either within or outside the network, exact or approximate.
Finally, symbolic meaning can serve to improve performance of S1. In other words, symbols which have been learned, derived or even invented can act as constraints on the large network and help improve learning performance as part of a positive cycle of learning and reasoning. Constraint satisfaction as part of the interplay between learning and reasoning is therefore another ingredient.

\textbf{Summary and Future Directions:} With the above five ingredients of neurosymbolic AI - gradient-based optimization, modularity, symbolic language, reasoning and constraint satisfaction - the reader will not be surprised to know that there are many outstanding challenges for neurosymbolic AI. First, no agreement exists on the best way of achieving the above combination of language and structure, of knowledge acquired by agents acting in an environment and the corresponding reasoning that an agent must implement to achieve its goals. It is highly desirable, though, that the study of how to achieve the combination of (symbolic) language and (neural) structure be principled, in that both language and structure should be formally specified with theorems proven about their correspondence or lack thereof. As done in the case of Noam Chomsky's language hierarchy, proofs are needed of the capability of different neural architectures at representing various logical languages. Proofs of correspondence have been shown between neural networks and propositional, nonmonotonic, modal, epistemic and temporal logic in \cite{Garcez_2008}. Similar proofs are required for first-order and higher-order logic.

Henry Kautz spoke of 6 types of neurosymbolic AI and said that what is important next is to work out which specific technique is best \cite{HenryKautz}. Kautz made a distinction between expert knowledge and commonsense knowledge and noted that one should not necessarily want to \emph{backpropagate through expert knowledge}. In this case, exact reasoning is needed \emph{à la} neural-symbolic computing with knowledge extraction. An equally valid argument exists for differentiable reasoning in the case of commonsense knowledge. The use of probabilistic languages or higher-order (functional or logical) languages may also have a central role in the technical debate, including on the best place for probability theory: in S1 or S2 or both? In the meantime, we say: translate back and forth between representations, take a principled approach, adopt a language as a constraint on the structure, seek to provide explanations, combine reasoning and learning, and repeat. 

We also set out three immediate challenges for neurosymbolic AI, each capable of spinning out multiple research strands which may become area defining in the next decade:

\begin{itemize}
\item Challenge 1:
First-order logic and higher-order knowledge extraction from very large networks that is provably sound and yet efficient, explains the entire model and local network interactions and accounts for different levels of abstraction.

\item Challenge 2:
Goal-directed commonsense and efficient combinatorial reasoning about what has been learned by a complex deep network trained on large amounts of multimodal data.

\item Challenge 3:
Human-network communication as part of a multi-agent system that promotes communication and argumentation protocols between the user and an agent that can ask questions and check her understanding.
\end{itemize}

Whether or not an AI system truly \emph{understands} what it does is another recurring theme in the current debate. A point made recently by Geoff Hinton on this issue was that: ``the goal posts keep changing. Before, if an AI system could translate a text, it would have been deemed as having understood the text, or if the system could sustain a conversation or describe the scene on an image. Now, none of these count as proper understanding''. One may argue that it is the very definition of AI and the recent success of deep learning itself that have been responsible for this situation. Perhaps, instead of proper ``understanding'', a more forgiving approach might be to specify comprehensibility tests of the kind used in schools to evaluate the performance of students on various subjects, including e.g. foreign language comprehensibility tests, as proposed by Stephen Muggleton. Neurosymbolic AI is in need of standard benchmarks and associated comprehensibility tests which could in a principled way offer a fair comparative evaluation with other approaches with a focus on learning from fewer data, reasoning about extrapolation, computational complexity and energy consumption. Just as the field of AI progressed when challenging applications were set such as chess playing, robotic football, self-driving vehicles and protein folding, neurosymbolic AI should benefit from a similar challenge and benchmark being set by the AI community specifically for the next decade.

\bibliographystyle{plain}

\end{document}